\documentclass[runningheads]{llncs}

\usepackage[algo2e,ruled,vlined]{algorithm2e}
\usepackage{amsmath, amssymb}
\usepackage{bbding}
\usepackage{diagbox}
\usepackage{float}
\usepackage{graphicx}
\usepackage{makecell}
\usepackage{multirow}
\usepackage{subfiles}

\begin{document}

\title{
    Learn to Predict Vertical Track Irregularity with Extremely Imbalanced Data
    \thanks{Supported By Science and Technology Research and Development Plan of China National Railway Group Co., Ltd. (P2018G051).}
}
\titlerunning{Learn to Predict Vertical Track Irregularity}

\author{
    Yutao Chen\inst{1}$^($\Envelope$^)$\orcidID{0000-0001-8826-2520} \and
    Yu Zhang\inst{2} \and
    Fei Yang\inst{2}
}
\authorrunning{Y. Chen et al.}

\institute{
    School of Computer Science (National Pilot Software Engineering School)\\
    Beijing University of Posts and Telecommunications, China\\
    \email{chnyutao@outlook.com}
    \and
    Infrastructure Inspection Research Institute\\
    China Academy of Railway Sciences Corporation Limited\\
    \email{\{zhangyu2016, yf2009\}@rails.cn}
}

\maketitle

\begin{abstract}
Railway systems require regular manual maintenance, a large part of which is dedicated to inspecting track deformation. Such deformation might severely impact trains’ runtime security, whereas such inspections remain costly for both finance and human resources. Therefore, a more precise and efficient approach to detect railway track deformation is in urgent need. In this paper, we showcase an application framework for predicting vertical track irregularity, based on a real-world, large-scale dataset produced by several operating railways in China. We have conducted extensive experiments on various machine learning \& ensemble learning algorithms in an effort to maximize the model's capability in capturing any irregularity. We also proposed a novel approach for handling imbalanced data in multivariate time series prediction tasks with adaptive data sampling and penalized loss. Such an approach has proven to reduce models' sensitivity to the imbalanced target domain, thus improving its performance in predicting rare extreme values.
\end{abstract}

\section{Introduction}

With the recent development in sensor technology, it has become easier to monitor railway infrastructures by deploying sensors on the track and trains. The installation of a massive amount of sensors in railway facilities has created an influx of data, which in turn spurs the development of more complicated information systems for monitoring railway conditions. Such systems could utilize the collected data for sophisticated maintenance scheduling and deterioration prevention, thus easing the burdens on finance and human resources.

In this paper, we focus on a crucial aspect of railway conditions -- vertical track irregularity. Vertical track irregularity has a decisive impact on trains' runtime security and stability. A slight, even imperceptible arch or depression in the track could lead to catastrophic results for a train running at high speed. However, massive, repeated manual gauging is still required to detect and fix emerging track irregularities to date.

Therefore, we proposed an application framework by using time series prediction techniques to detect vertical track irregularity. The prediction is carried out in two dimensions:
\begin{itemize}
    \item \textit{Spatial dimension}: we utilize data from adjacent track points to predict the condition of the current track point;
    \item \textit{Temporal dimension}: we utilized historical data of the current track point to predict its condition at this moment.
\end{itemize}
The condition of a point on the track is primarily measured by its track height. Abnormally high track heights might indicate an arching deformation in adjacent areas, while low track heights might indicate a depressed deformation. Such a framework could be applied in real-life engineering cases, such as preventive maintenance.

We have seen relevant works on applying machine learning techniques to railway maintenance. Tsunashima\cite{tsunashima2019condition} uses a support vector machine to classify car-body vibration signals in order to determine track conditions. Gibert\cite{gibert2016deep} uses deep convolutional neural networks\cite{krizhevsky2012imagenet} to inspect railway track conditions through photos. Krummenacher\cite{krummenacher2017wheel} uses time series prediction to determine wheel defects, and Flink\cite{fink2013predicting} uses time series prediction to predict railway speed restrictions.

Our contribution is this paper could be phrased as follows:
\begin{itemize}
    \item To the authors' knowledge, we are among the first to apply time series prediction techniques, especially recurrent neural networks, to track irregularity monitoring. Our experience and solution could be a viable reference for the community;
    \item We proposed a novel approach, combining adaptive data sampling and penalized loss, to fight data imbalance in multivariate time series regression tasks. This approach is proved to be effective and should be a complement to existing imbalanced learning algorithms;
    \item We have conducted exhaustive experiments on various time series prediction algorithms, including ARIMAX\cite{williams2001multivariate}, LSTM\cite{hochreiter1997long}, GRU\cite{chung2014empirical}, and CNN\cite{krizhevsky2012imagenet}. We have also adopted multiple ensemble learning methods, including Bagging, Boosting, and Stacking, for optimizing the model's performance.
\end{itemize}

\section{Methodology}

In this section, we will discuss the algorithms vital to the development of this paper.

\subsection{Time Series Prediction}

\subsubsection{ARIMAX\cite{williams2001multivariate}} is an extension of the autoregressive integrated moving average (ARIMA) model. ARIMAX accepts exogenous variables, which enables multivariate time series regression.

\subsubsection{LSTM\cite{hochreiter1997long}}
\begin{figure}[h]
    \centering
    \includegraphics[width=0.5\textwidth]{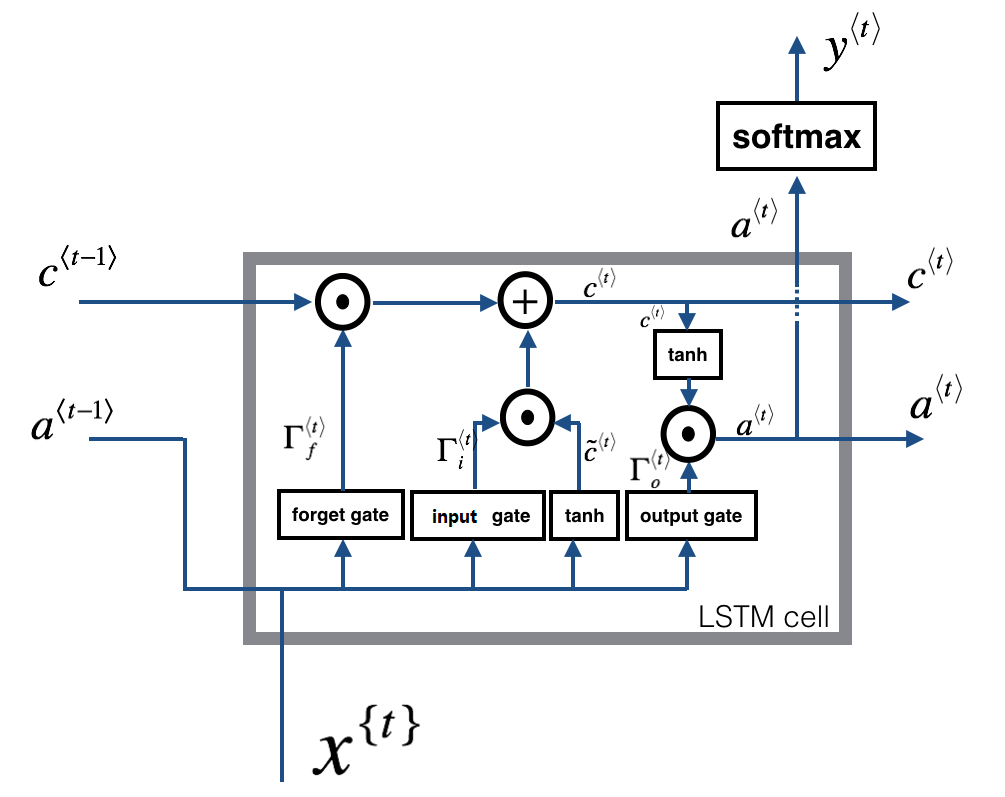}
    \caption{The internal structure of an LSTM memory unit.}
    \label{fig:lstm-cell}
\end{figure}

, short for Long Short-Term Memory, is a variation of the recurrent neural network. It introduces a memory cell, as shown in Fig.\ref{fig:lstm-cell}, with an update gate and an output gate. These gates hold the memory cell's internal states and help improve the network's memory over long sequences.

\subsubsection{GRU\cite{chung2014empirical}}, short for Gated Recurrent Unit, is a simplification over LSTM. It removes the output gate in an effort to reduce parameters and accelerate computation. Experiments have demonstrated that GRU can achieve similar or better performance compared to that of LSTM, even with fewer parameters.

\subsubsection{CNN\cite{krizhevsky2012imagenet}}, short of convolutional neural networks, is typically seen in computer vision tasks. However, it could also be adapted for time series processing\cite{zhao2017convolutional} by viewing the multivariate time series as a matrix, as we do with the pictures. For example, given an input multivariate time series of shape $n \times m$ (where $m$ is the number of features and $n$ is the length of the time series ), a convolutional kernel of $k \times m$ (where $k$ is the size of kernel and $k<l$) is applied along the feature axis and returns an output vector of $(n-k+1) \times 1$. More kernels can be used in order to extract more features from the input time series.

\subsection{Ensemble Learning}

Apart from multivariate time series regression, we use ensemble learning methods for further optimizing the model's performance.

\subsubsection{Bagging\cite{zhou2009ensemble}}, short for boostrap aggregating, is an ensemble learning method in which we train $m$ base learners \textit{parallelly} on $m$ different datasets. The $m$ datasets, given a original dataset $D$ of size $n$, is sampled from $D$ uniformly and with replacement. All $m$ datasets share the same size $n'$, where $n' < n$.

\subsubsection{Boosting\cite{zhou2009ensemble}} is not algorithmically constrained, but most boosting algorithms consist of iteratively learning weak learners with respect to a distribution and adding them to a final strong learner. The Adaboost algorithm is probably the most popular derivation in the boosting family. Unlike bagging, the base learners in boosting are required to be trained \textit{sequentially} as the next base learner is logically dependent on the performance of the previous base learner.

\subsubsection{Stacking}
With ensemble learning methods like \textit{Bagging} and \textit{Boosting}, we train a set of base learners. These base learners are later aggregated usually using majority voting (for classification) or weighted averaging (for regression). \textit{Stacking}\cite{zhou2009ensemble} allows more complex aggregation by introducing another learner, which could be linear regression or even multi-layer perceptron, on top of the base learners. The stacked learner takes the outputs of base learners as inputs and then produces a final output. The stacked learner could be trained and fine-tuned individually.

\subsection{Imbalanced Regression}

In a normally operating railway, vertical track irregularities are rarely seen -- otherwise the railway won't be operating normally. In our case, such irregularities take up less than 0.1\% of all the data. In this case, traditional regression algorithms with MSE/RMSE losses would barely work, hence we need algorithms and metrics specificly designed for machine learning with imbalanced data.

\subsubsection{Utility-based Regression\cite{torgo2007utility}}

In utility-based regression, each target variable $y$ is assigned with a relevance value $\Phi(y)$, indicating its importance. Rarer target values have higher relevance. When calculating the loss, these relevance values are taken into account by multiplying $\Phi(y)\times\Phi(\hat{y})\times loss$, so that the model will be penalized with a higher loss if it made a wrong prediction on rarer values.

\subsubsection{SmoteR\cite{torgo2013smote}}
SmoteR is short for Smote for Regression. Smote is originally an algorithm for handling classification problems with imbalanced data. SmoteR uses under-sampling for the majority so that common data will be reduced, while using over-sampling for the minority by generating synthetic cases so that rare data will be amplified.

\subsubsection{Precision, Recall, and F-Score}
As extreme target values are so rarely seen, loss functions like MSE/RMSE won't be able to reflect the model's capability in predicting extreme values. Instead, we should use precision, recall, and the F-score.
$$
\begin{aligned}
    &\text{precision} = \dfrac{\text{tp}}{\text{tp+fp}}\\
    &\text{recall} = \dfrac{\text{tp}}{\text{tp+fn}}\\
    &\text{F}_1 = 2\times\dfrac{\text{precision}\times\text{recall}}{\text{precision}+\text{recall}}\\
\end{aligned}
$$
Here, \textit{tp} denotes numbers of true positive cases, \textit{fp} denotes number of false positive cases, and \textit{fn} denotes number of false negative cases.

\section{Task and Data}

Our dataset is collected from the Beijing-Tianjin railway line in China every month from 2011 to 2017. The dataset is separated by month and distributed in multiple CSV files. Each CSV file contains approximately 420,000 lines.

Each line represents the condition of a specific point on the track at the time of data collection, and could be uniquely identified by two variables: \textit{mileage} and \textit{meters}. The former specifies one kilometer of track along the railway line, while the latter specifies a unique point within the kilometer. Since the sampling data points distribute uniformly along the line with a constant interval of 0.25m, the value of \textit{meters} variable ranges from 0.0, 0.25, 0.5, 0.75 to 999.75.

The data initially contains 32 columns, apart from the two identifier columns: \textit{mileage} and \textit{meters}. Two of these columns, namely \textit{left/right track height}, are the target variables we want to predict. The other 30 columns are therefore feature columns, including \textit{track gauge}, \textit{vertical/horizontal acceleration}, etc. Most of these feature columns could be removed at train time, however, because:
\begin{itemize}
    \item Some of these feature columns (\textit{flags}, \textit{event}, \textit{speed}, etc.) are invalid as all cells in the column share the same value, for instance, zero. These columns are reserved for sensors yet to be installed.
    \item Others of these feature columns have very little correlation with the target we want to predict, namely \textit{left/right track height}.
\end{itemize}

\begin{figure}[t]
    \centering
    \includegraphics[width=0.5\textwidth]{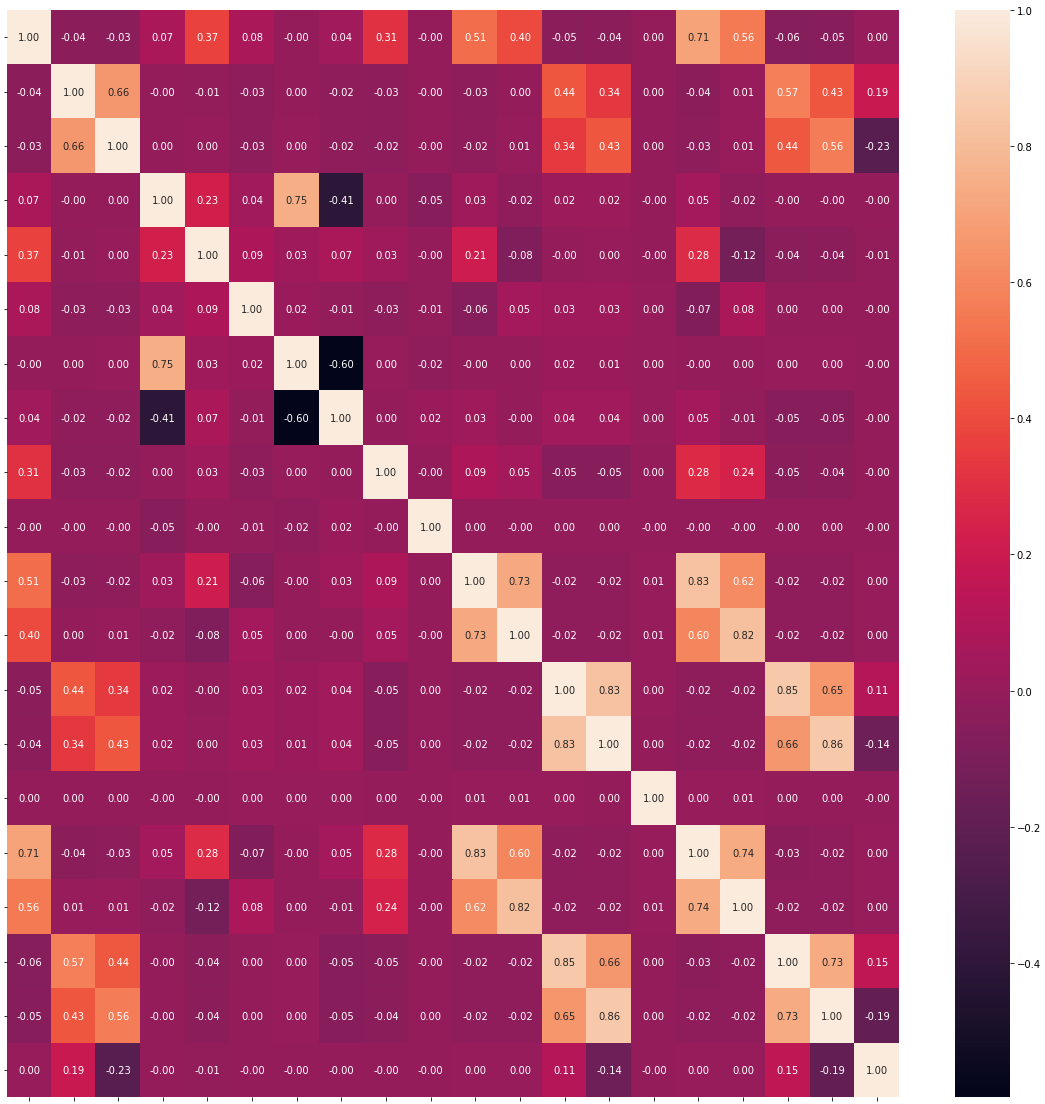}
    \caption{The Spearman correlation coefficient matrix.}
    \label{fig:spearman}
\end{figure}

By using the Spearman correlation coefficient, we could compute a correlation coefficient matrix, as shown in Fig.\ref{fig:spearman}, representing the correlation between any two columns in the dataset.
Here we prefer the Spearman coefficient over the Pearson coefficient for evaluating correlation, as the Spearman coefficient could capture non-linear correlations. We remove columns that are less correlated with the target variables (\textit{left/right track height}), and end up with only 10 feature columns. Experiments have also proven that the model could attain similar performance even with less than half of the original features.

The dataset is then transformed into spatial and temporal time series using a sliding window. For spatial time series generation, we sample along the track, spaced every 0.25m.The window width is 8.
$$
\begin{bmatrix}
    x + 0.00\\
    x + 0.25\\
    x + 0.50\\
    x + 0.75\\
    x + 1.00\\
    x + 1.25\\
    x + 1.50\\
    x + 1.75\\
\end{bmatrix}
\qquad
\begin{bmatrix}
    x + 0.25\\
    x + 0.50\\
    x + 0.75\\
    x + 1.00\\
    x + 1.25\\
    x + 1.50\\
    x + 1.75\\
    x + 2.00\\
\end{bmatrix}
\qquad
\begin{bmatrix}
    x + 0.50\\
    x + 0.75\\
    x + 1.00\\
    x + 1.25\\
    x + 1.50\\
    x + 1.75\\
    x + 2.00\\
    x + 2.25\\
\end{bmatrix}
\qquad\dots
$$
For temporal time series generation, we sample every year, at the same month and track point. The window width is 4.
$$
\begin{bmatrix}
    April\ 2011\\
    April\ 2012\\
    April\ 2013\\
    April\ 2014\\
\end{bmatrix}
\qquad
\begin{bmatrix}
    April\ 2012\\
    April\ 2013\\
    April\ 2014\\
    April\ 2015\\
\end{bmatrix}
\qquad
\begin{bmatrix}
    April\ 2013\\
    April\ 2014\\
    April\ 2016\\
    April\ 2016\\
\end{bmatrix}
\qquad\dots
$$

The task now could be formulated as finding such a function $f:\ \mathbb{R}^{(n \times m)}\to\mathbb{R}$, where $n$ is the length of input time series (8 for spatial time series, 4 for temporal time series) and $m$ is the number of features (10 after we apply the Spearman coefficient to remove features).

\section{Proposed Solution}

\subsection{Spatial Time Series Prediction}

For spatial time series prediction, five algorithms are explored and compared in this paper:
\begin{itemize}
    \item \textit{Linear Regression} serves as a base line for other models with sequential prediction capabilities. It ignores the temporal information and use the other nine exogenous features as inputs to predict the target track height, with mean squared error as the error function.
    \item \textit{ARIMAX} is a machine learning approach for time series prediction. We tried several parameter combinations including $(p=3,\ d=0,\ q=0)$, $(p=5,\ d=1,\ q=0)$, and $(p=8,\ d=2,\ q=3)$. The one with the best performance is promoted to compete with neural networks.
\end{itemize}

As for deep learning algorithms, LSTM, GRU, and CNN are involved:
\begin{itemize}
    \item \textit{Long Short Term Memory} is an outstanding variant of RNN for time series analysis. In this paper, we use one layer of LSTM cells for inputs, along with one neuron densely connected to the previous layer for output.
    \item \textit{Gated Recurrent Unit} is similar to LSTM but without an output gate. In this paper, we use one layer of GRU cells for inputs, along with one neuron densely connected to the previous layer for output.
    \item \textit{Convolutional Neural Network} can be used for time series forecasting as well. In this paper, we use a convolution layer with five kernels of size 5 for inputs, along with one neuron densely connected to the previous layer for output.
\end{itemize}

Note that for all neural networks, the Adam optimization algorithm1\cite{kingma2014adam} is used. The batch size is 128 and the loss function is mean squared error. Two mechanisms exist as prevention against overfitting:
\begin{itemize}
    \item All input layers (LSTM, GRU, CNN) are restrained by a L2 regularizer;
    \item A validation set is used for early stopping with a three degree patience. If the loss on the validation set has been rising for at least three epochs, the training process will be stopped and the model's parameters will be restored to that of three epochs ago.
\end{itemize}

Spatial time series prediction itself might seem less intuitive than temporal time series prediction, as the current condition of a specific railway track point could be directly read from the sensors. However, spatial time series prediction could still be helpful for locating hardware failures in sensors, for example, when the some sensor emits data that dramatically deviates from the model's prediction.

\subsection{Temporal Time Series Prediction}

For temporal time series prediction, we employed ensemble learning methods with LSTM, GRU, and CNN as base learning algorithms. We use:
\begin{itemize}
    \item \textit{bagging} to generate $m$ train sets and parallelly produce $m$ base learners on the $m$ train sets;
    \item \textit{boosting} to train $m$ base learners sequentially and adjust the train set accordingly every iteration.
\end{itemize}
The algorithm for boosting is described in Algorithm \ref{alg:boosting}. Finally, a logistics regression is used for \textit{stacking} the $m$ base learners, produced by either bagging or boosting.

\begin{algorithm2e}[h]
    \caption{A simplified boosting algorithm.}
    \label{alg:boosting}
    \SetAlgoLined
    Let $X_1=<x_1,x_2,\dots,x_n>$, $Y_1=<y_1,y_2,\dots,y_n>$\\
    \For{i = 1 to m}{
        train a base learner $L_i$ on $(X_i,Y_i)$\\
        \For{j = 1 to i}{
            calculate the absolute bias $E=|Y-L_i(X)|$\\
            \For{k = 1 to n}{
                \If{$E_k > threshold$}{
                    add $x_k$ to $X_{i+1}$\\
                    add $y_k$ to $Y_{i+1}$
                }
            }
        }
    }
\end{algorithm2e}

\subsection{Fighting Data Imbalance}

\begin{figure}[t]
    \centering
    \minipage{0.5\textwidth}
        \includegraphics[width=\linewidth]{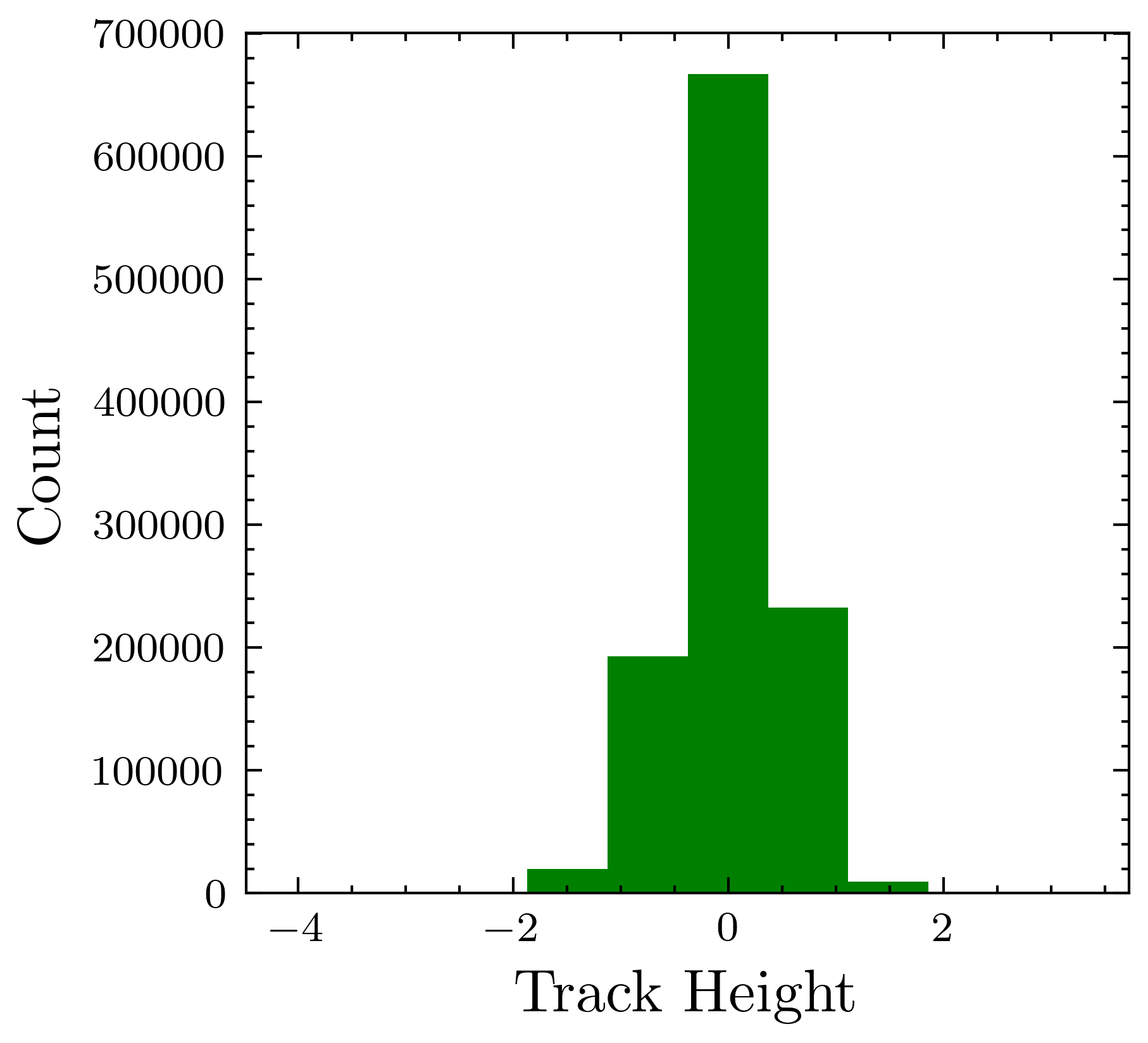}
        \caption{Track height.}
        \label{fig:height-dist}
    \endminipage
    \minipage{0.5\textwidth}
        \includegraphics[width=\linewidth]{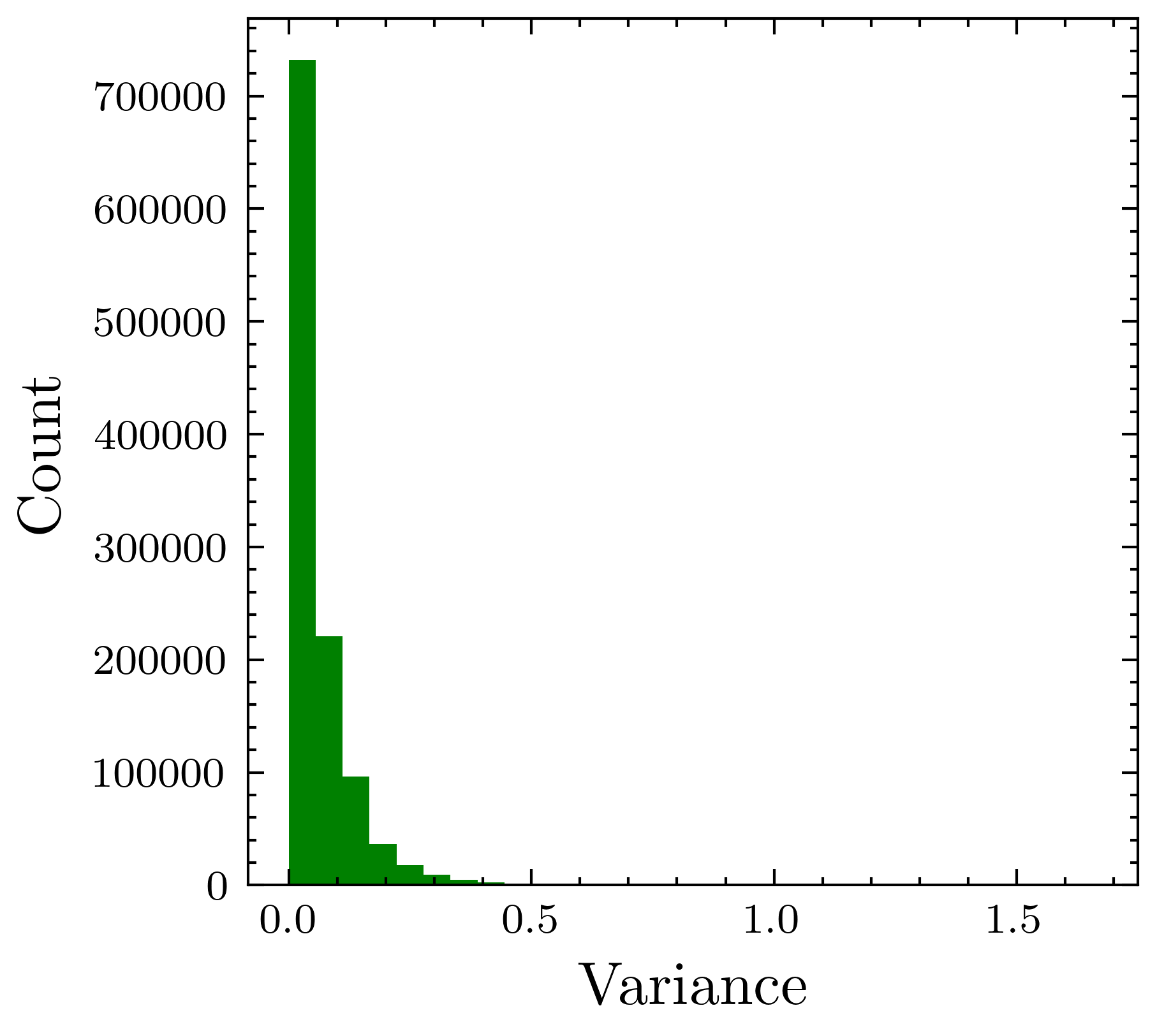}
        \caption{Spatial time series variance.}
        \label{fig:variance-dist}
    \endminipage
\end{figure}

A serious issue yet to be resolved is that our dataset is extremely imbalanced. Fig.\ref{fig:height-dist} shows that the majority of track heights reside within $[-1, 1]$. Such imbalance could invalidate our previous effort to minimized the MSE loss. Hence, we proposed a novel approach for handling imbalanced multivariate time series regression with adaptive data sampling and penalized loss.

We first measure the variance of track heights within one time series. As shown in Fig.\ref{fig:variance-dist}, most time series have very low variance, indicating the stability of this track segment. In constrast, those time series with high variances are much more liable to produce vertical track irregularities. We consider time series with a variance less than the \textit{threshold} (for instance 1.15) \textit{common} and otherwise \textit{rare}. Inspired by SmoteR\cite{torgo2013smote}, we use random under-sampling to reduce common data, while over-sampling the rare data by replicating samples from them with boostrapping.

It could be tricky to determine an appropriate proportion for under-sampling and over-sampling -- how much common data should we discard and how much rare data should we replicate? We decide upon an adaptive approach in which several predefined proportions are engaged in a one-epoch training. The result of the one-epoch training are then evaluated to help us determine a best sampling proportion.

We also adopted a form of penalized mean squared error as the loss function, inspired by utility-based regression\cite{torgo2007utility}. Here is the penalty matrix, where $0 \leq p_4 < p_1 < p_2 < p_3$:
\begin{table}[H]
    \centering
    \setlength{\tabcolsep}{5pt}
    \begin{tabular}{|c|c|c|}
        \hline
        \diagbox{predicted}{correct} & common & rare\\
        \hline
        common & $p_1$ & $p_2$\\
        \hline
        rare & $p_3$ & $p_4$\\
        \hline
    \end{tabular}
\end{table}
With this penalty matrix, the mean squared error could be rewritten as:
$$
\begin{aligned}
    \text{MSE}=\dfrac{1}{n}\sum\limits_{i=1}^{n}{[(y_i - \hat{y}_i)^2\times\Phi(y_i, \hat{y}_i)]}
\end{aligned}
$$
where $y_i$ is the correct value, $\hat{y}_i$ is the predicted value, and $\Phi(y_i, \hat{y}_i)$ is their corresponding term in the penalty matrix.

In this paper, we use $p_1=1, p_2=3, p_3=10, p_4=0.8$.

\setlength{\tabcolsep}{10pt}
\section{Performance Evaluation}

\subsection{Time Series Prediction}

In Table \ref{tab:performance-ml}, the performances of several machine learning algorithms are exhibited. ARIMAX has slight performance improvement over that of linear regression.

\begin{table}[H]
    \centering
    \caption{Performance of machine learning algorithms.}
    \begin{tabular}{c|cc|cc}
        \hline
        \multirow{2}{4em}{Model} & \multicolumn{2}{c|}{Train} & \multicolumn{2}{c}{Test} \\
        & MSE & MAE & MSE & MAE \\
        \Xhline{2\arrayrulewidth}
        Linear Regression & 0.1826 & 0.3079 & 0.1754 & 0.3052 \\
        ARIMAX(3,0,0) & 0.1601 & 0.3045 & 0.1419 & 0.2766 \\
        ARIMAX(5,1,0) & 0.1427 & 0.2685 & 0.1406 & 0.2611 \\
        ARIMAX(8,2,3) & 0.1379 & 0.2576 & 0.1317 & 0.2438 \\
        \hline
    \end{tabular}
    \label{tab:performance-ml}
\end{table}

In Table \ref{tab:performance-dl}, the performances of  deep learning algorithms are exhibited. Experiments have been repeated for 3 times with randomized parameter initialization and then averaged to improve credibility. LSTM and GRU have achieved very similar results, whereas CNN usually has higher MSE but lower MAE.

\begin{table}[h]
    \centering
    \caption{Performance of deep learning algorithms.}
    \begin{tabular}{c|cc|cc|cc}
        \hline
        \multirow{2}{4em}{Mode} & \multicolumn{2}{c|}{Train} & \multicolumn{2}{c|}{Val} & \multicolumn{2}{c}{Test} \\
        & MSE & MAE & MSE & MAE & MSE & MAE \\
        \Xhline{2\arrayrulewidth}
        LSTM & 0.0495 & 0.1651 & 0.0496 & 0.1659 & 0.0499 & 0.1652 \\
        GRU & 0.0499 & 0.1658 & 0.0501 & 0.1667 & 0.0502 & 0.1659 \\
        CNN & 0.0517 & 0.1628 & 0.0518 & 0.1635 & 0.0521 & 0.1629 \\
        \hline
    \end{tabular}
    \label{tab:performance-dl}
\end{table}

\subsection{Ensemble Learning}

In Table \ref{tab:performance-el}, the performances of ensemble learning algorithms are exhibited. Bagging has the best performance, followed by boosting and tailed by no ensemble learning.

\begin{table}[H]
    \centering
    \caption{Performance of ensemble learning algorithms.}
    \begin{tabular}{c|cc|cc|cc}
        \hline
        \multirow{2}{4em}{Model} & \multicolumn{2}{c|}{None} & \multicolumn{2}{c|}{Bagging} & \multicolumn{2}{c}{Boosting} \\
        & MSE & MAE & MSE & MAE & MSE & MAE \\
        \Xhline{2\arrayrulewidth}
        LSTM & 0.0473 & 0.1648 & 0.0423 & 0.1607 & 0.0458 & 0.1623\\
        GRU & 0.0483 & 0.1647 & 0.0431 & 0.1611 & 0.0467 & 0.1628 \\
        CNN & 0.0551 & 0.1621 & 0.0519 & 0.1597 & 0.0569 & 0.1603 \\
        \hline
    \end{tabular}
    \label{tab:performance-el}
\end{table}

\subsection{Imbalanced Regression}

For evaluating the model's capability in capturing irregularity, we consider time series with a target track height within $[-1, 1]$ negative (common/regular) cases, otherwise positive (rare/irregular) cases. In this way, the regression task is reduced to a binary classification problem, where we could apply precision, recall, and F-score to evaluate the model's capability in capturing irregularity.

\begin{table}[H]
    \centering
    \caption{Performance of difference imbalanced learning algorithms}
    \begin{tabular}{c|ccccc}
        \hline
        Algorithm & MSE & MAE & Precision & Recall & F-Score\\
        \Xhline{2\arrayrulewidth}
        None & 0.0471 & 0.1627 & 0.55 & 0.34 & 0.4202\\
        Adaptive Sampling & 0.0512 & 0.1873 & 0.59 & 0.57 & 0.5798\\
        Penalized Loss & & & 0.64 & 0.72 & 0.6776\\
        Both & & & 0.63 & 0.79 & 0.7010\\
        \hline
    \end{tabular}
    \label{tab:performance-im}
\end{table}

As shown in Table \ref{tab:performance-im}, when both adaptive sampling and penalized loss are applied, 79\% of the vertical track irregularities can be captured.

The base model for time series regression used here is LSTM, with the same architecture and setup as described in the previous section.

\section{Conclusion}

In this paper, we showcase a application framework for predicting vertical track irregularity with a extremely imbalanced dataset, by using time series prediction techniques. We compared various algorithms for time series prediction and reached at the conclusion that neural networks generally outperforms traditional time series forecasting algorithms like ARIMAX. We also proposed a novel approach to fight data imbalance in multivariate time series regression problems by combining adaptive sampling and penalized loss.

In practice, it is recommended to use LSTM/GRU with bagging for best performance, as well as adaptive data sampling and penalized loss for averting imbalanced regression pitfalls.

\bibliographystyle{splncs04}
\bibliography{reference}

\end{document}